\documentclass[letterpaper, 10 pt, conference]{ieeeconf}  

\IEEEoverridecommandlockouts                              

\usepackage{amsmath}       
\usepackage{amssymb}       
\usepackage{graphicx}      
\usepackage{float}

\usepackage{algorithm}     
\usepackage[noend]{algpseudocode} 
\usepackage{booktabs}      
\usepackage{cleveref}      
\usepackage{xcolor}        
\usepackage{graphicx}
\usepackage{subcaption} 
\usepackage{cite}
\usepackage{empheq}

\renewcommand{\baselinestretch}{0.972}

\newtheorem{theorem}{Theorem}
\title{\LARGE \bf
Behaviorally Adaptive Multi-Robot Hazard Localization in Failure-Prone, Communication-Denied Environments}

 \author{
    Alkesh K. Srivastava$^{1}$, Aamodh Suresh$^{2}$, and Carlos Nieto-Granda$^{2}$%
    \thanks{$^{1}$Alkesh K. Srivastava is with the Department of Mechanical Engineering, Temple University, Philadelphia, PA, USA. He is also a research intern at the U.S. Army Research Laboratory (ARL), Adelphi, MD, USA. {\tt\small alkesh@temple.edu}}%
    \thanks{$^{2}$Aamodh Suresh and Carlos Nieto-Granda are with the U.S. DEVCOM Army Research Laboratory (ARL), Adelphi, MD, USA. {\tt\small aamodh@gmail.com}, {\tt\small carlos.p.nieto2.civ@army.mil}}%
}

\begin{document}

\maketitle
\thispagestyle{empty}
\pagestyle{empty}

\begin{abstract}
We address the challenge of multi-robot autonomous hazard mapping in high-risk, failure-prone, communication-denied environments such as post-disaster zones, underground mines, caves, and planetary surfaces. In these missions, robots must explore and map hazards while minimizing the risk of failure due to environmental threats or hardware limitations. We introduce a behavior-adaptive, information-theoretic planning framework for multi-robot teams grounded in the concept of Behavioral Entropy (BE), that generalizes Shannon entropy (SE) to capture diverse human-like uncertainty evaluations.
Building on this formulation, we propose the Behavior-Adaptive Path Planning (BAPP) framework, which modulates information gathering strategies via a tunable risk-sensitivity parameter, and present two planning algorithms: BAPP-TID for intelligent triggering of high-fidelity robots, and BAPP-SIG for safe deployment under high risk. We provide theoretical insights on the informativeness of the proposed BAPP framework and validate its effectiveness through both single-robot and multi-robot simulations. 
Our results show that the BAPP stack consistently outperforms Shannon-based and random strategies: BAPP-TID accelerate entropy reduction, while BAPP-SIG improves robot survivability with minimal loss in information gain. In multi-agent deployments, BAPP scales effectively through spatial partitioning, mobile base relocation, and role-aware heterogeneity. These findings underscore the value of behavior-adaptive planning for robust, risk-sensitive exploration in complex, failure-prone environments.

\end{abstract}

\section{Introduction}

\begin{figure}[t]
    \centering
    \centering
        \includegraphics[width=\linewidth]{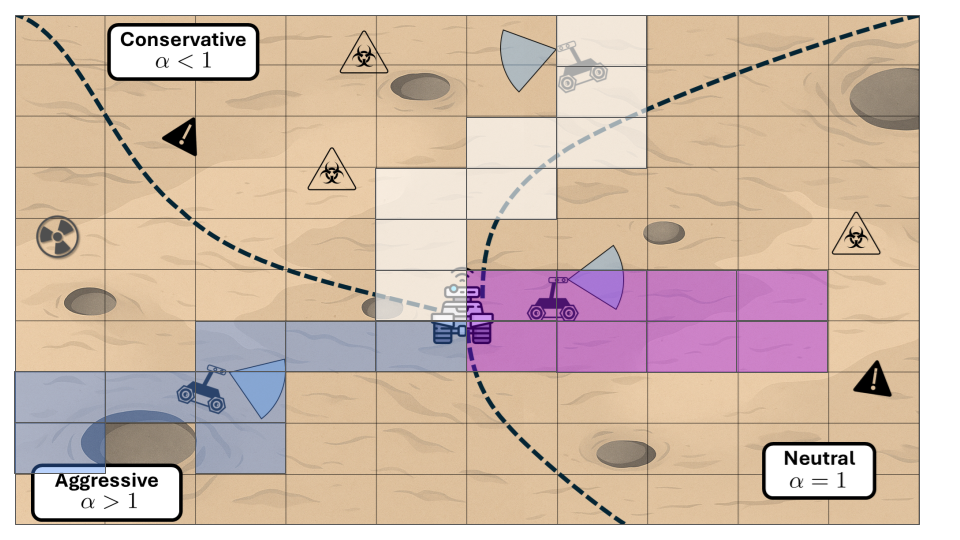}
    \caption{\small \textit{Motivating example of our proposed framework in hazardous planetary environments.} A large expensive mobile base robot, tasked with locating hazards, deploys small, inexpensive agents into partitioned sectors (using current hazard belief map) to follow a planned path (highlighted squares). Each agent is assigned an appropriate behavior mode: \textit{conservative} avoids threats, \textit{neutral} explores uniformly, and \textit{aggressive} targets high-uncertainty regions despite nearby hazards. Success or failure of each deployment is treated as a latent observation of hazards on the path and fed back to update the hazard belief map.
    }
    \vspace{-15pt}

    \label{fig:concept-figure}
\end{figure}

Robotic exploration of unknown, hazardous environments poses significant challenges—especially when missions are autonomous, communication-denied, and failure-prone. Such conditions arise in search and rescue after natural disasters, underground mine inspections, planetary exploration, and wildfire surveillance. In these scenarios, human intervention is infeasible, requiring robots to navigate uncertain terrain, gather critical information, and survive long enough to relay it, often without centralized control or reliable communication. As illustrated in~\Cref{fig:concept-figure}, scalable deployment in such environments often requires coordinating heterogeneous agents with varying levels of durability and risk tolerance. These constraints pose a fundamental question: how can a team of robots plan exploration paths that are both information-rich and risk-sensitive?

Classical approaches address this challenge using information-theoretic planning, where exploration decisions are guided by maximizing expected information gain based on measures like Shannon entropy (SE)~\cite{Shannon.BellJ.1948}. Seminal efforts have demonstrated how robots can efficiently reduce uncertainty through entropy-driven planning, whether in UAV-based target search~\cite{Furukawa.etal.ICRA.2006, Tisdale.etal.RAM.2009}, distributed sensor networks~\cite{Julian.etal.IJRR.2012, Khan.etal.TCNS.2015}, hazard-aware multi-robot deployments~\cite{schwager2017multi}, or terrain monitoring~\cite{popovic2020informative}. 
Additional research has investigated semantic-aware planning~\cite{ginting2023safe}, SLAM in GPS-denied environments~\cite{ebadi2022present}, and belief-space planning under uncertainty~\cite{agha2018slap}. In parallel, vision-based localization~\cite{nath2022drone, goforth2019gps} and aerial damage assessment~\cite{greenwood2020flying} have demonstrated the growing demand for autonomy in hazardous environments. Recent efforts in communication-denied settings~\cite{otte2021path, srivastava2022distributed, srivastava2023path} have adapted these principles to path-based exploration. However, many of these approaches still assume fixed robot behavior~\cite{otte2021path, srivastava2023path}, static base stations~\cite{srivastava2023path, mendelsohn2024enhancing, mcguire2024valuing}, or homogeneous teams~\cite{srivastava2022distributed}, which limit their applicability in real-world scenarios.  

Multi-agent coordination strategies have tackled cooperative exploration~\cite{bourgault2004decentralized, Julian.etal.IJRR.2012, hollinger2014distributed}, multi-target tracking~\cite{dames2015autonomous}, and hazard-aware control~\cite{schwager2017multi}. Works such as~\cite{Chung.Burdick.TRO.2012, Berger.Jens.CEC.2010} emphasize probabilistic decision-making and constrained communication in multi-agent settings. Risk-aware exploration in subterranean GPS-denied environments~\cite{patel2024towards} and multi-modal disaster mapping~\cite{nath2022drone} further underscore the importance of behavior-adaptive, infrastructure-free autonomy. Nonetheless, most of these approaches assume uniform agent roles and do not explicitly reason about survivability, energy constraints, or behavioral adaptation under uncertainty.

To address these limitations, we introduce the \textit{Behavior-Adaptive Path Planning (BAPP)} framework—a modular, scalable approach for multi-robot exploration and hazard localization in failure-prone, communication-denied environments. BAPP integrates risk-sensitive decision-making, role-aware deployment, and mobile base relocation via behavior modulation using the tunable $\alpha$ parameter of \textit{Behavioral Entropy} (BE)~\cite{Suresh.etal.RAL.2024}. This enables safer information gathering, triggered deployment of high-fidelity agents, and entropy-aware base movement to improve coverage. BE generalizes Shannon entropy using the Prelec function~\cite{prelec1998probability} to adjust perceived uncertainty, allowing robots to trade off exploration and caution. BAPP treats each path’s outcome as an observation of latent risk, updating a spatial belief map via Bayesian inference. These updates guide adaptive triggering and risk-aware planning across deployments. While BE has seen use in single-agent exploration settings~\cite{Suresh.etal.RAL.2024}, BAPP extends it to distributed, heterogeneous teams under uncertainty and resource constraints.

\textbf{Our contributions are threefold:} First, we establish theoretical guarantees showing that BE always admits an $\alpha$ yielding greater or at least as much informativeness than SE in hazardous, communication-denied environments. Second, we validate behavior-modulated planning in failure-prone simulations, demonstrating consistent gains in both robustness and information acquisition. Third, we extend the framework to distributed multi-agent deployments, introducing mechanisms for spatial partitioning, and mobile base relocation that collectively enable scalable, risk-sensitive exploration.

\section{Problem Formulation}
We consider a heterogeneous multi-robot team tasked with mapping hazardous regions (e.g., fire, gas, radiation) in an unknown environment having \textit{failure-prone} and \textit{communication-denied} conditions. Such scenarios arise in planetary exploration, underground mine surveys, and post-disaster response, where real-time communication coordination is infeasible and robot survivability is limited. 

We consider the environment as a 2D compact set $\mathcal{E} \subset \mathbb{R}^2$, discretized into square grid cells \( \mathcal{E}_{\text{grid}} \in \mathbb{Z}^{a \times b} \). Each cell \( c_i \in \mathcal{E}_{\text{grid}} \) contains a latent binary variable \( X_i \in \{0, 1\} \), indicating whether the cell is hazardous (\( X_i = 1 \)) or safe (\( X_i = 0 \)). Additionally, to capture the severity of hazards, each hazardous cell (\( X_i = 1 \)) is associated with a lethality \(\lambda_i \in [0,1]\), which defines the probability that a robot fails upon entering it. For instance, a high-lethality cell may correspond to an unstable crater rim, an area of intense radiation, or a fire zone with reduced visibility.

Robots are deployed from a mobile base $R$ that periodically relocates to improve coverage and acts as a base of reconnaissance. Each deployed robot $r_j$ is assigned a fixed-length path relative to the base station \(\zeta = \langle c_1, c_2,\ldots, c_T\rangle\), composed of $T$ sequential cells reachable via 9-connected grid motion. Robots traverse their assigned paths autonomously and without communication. The only observable outcome is a binary variable \( \Theta(\zeta) \in \{0, 1\} \), where \( \Theta = 1 \) indicates mission failure and \( \Theta = 0 \) denotes a successful return. The robot fails along path \( \zeta \) if it enters any hazardous cell \( c_i \in \zeta \) and failure is realized (\( Z_i = 1 \)) for any \( c_i \in \zeta \). Otherwise, it returns to the base safely (\( Z_i = 0 \)). 

Because direct measurements of hazard presence are unavailable, uncertainty is modeled through the survivability outcome of each deployment. Specifically, when a robot visits a hazardous cell, the probability of mission failure is equal to the cell’s lethality $ \lambda_i $, which reflects how likely the environment is to confirm its hazardous nature. Conversely, even in safe cells, robots may still fail due to internal malfunction or external factors, modeled via a nominal malfunction rate $ \gamma_i $. This formulation implicitly relates survivability-based outcomes to belief updates, enabling probabilistic reasoning over latent hazard presence.

The base $R$ maintains a belief $p_i \in [0,1] $ for each cell $c_i \in \mathcal{E}_{\text{grid}}$, indicating whether it is hazardous ($X_i=1$) with $p_i=0.5$ implying no knowledge, $p_i=1.0$ implying hazard, and $p_i=0$ implying free space. The global hazard belief map is then \( \mathcal{B} = [p_1, \ldots, p_{ab}] \). Starting with no knowledge ($p_i=0.5, \forall i$), the goal is to infer the global hazard map \( \mathbf{X} \) through repeated deployments of a team of robots and updating $\mathcal{B}$ after receiving the observation $Z$.

To this end, we introduce a behavior-adaptive framework that modulates information gathering using a single tunable parameter. Inspired by human decision-making under uncertainty, our approach enables robots to adapt their behavior dynamically based on mission context or progress, offering a robust solution for mapping in failure-prone, communication-denied environments. 

\section{Information-Theoretic Framework}
\label{sec:behavioral-entropy}
In hazardous, communication-denied environments where robot failures are costly and sensing opportunities are limited, planning must carefully balance the objective of reducing uncertainty while ensuring survivability. Classical entropy-based approaches prioritize information gain but often neglect the risks associated with environmental hazards or hardware failure. To overcome this limitation, we introduce a behavior-adaptive, information-theoretic framework that explicitly incorporates variety of risk sensitive behaviors into planning. We outline the foundational principles of our approach next.

\paragraph*{Behavioral Entropy~\cite{Suresh.etal.RAL.2024,suttle2025behavioral}}

Suresh \textit{et al.}~\cite{Suresh.etal.RAL.2024} introduced Behavioral Entropy $H^B$ by composing the Boltzmann-Gibbs-Shannon (BGS) entropy with Prelec's weighting function $w(p)$~\cite{prelec1998probability}, originally developed to model human perception of uncertainties~\cite{dhami2016foundations}. The Prelec weighting function is given by \(w(p) = \exp({-\beta(-\log p)^\alpha})\),
where $\alpha,\beta > 0 $ control the shape and fixed points of the weighting $w$, leading to different perception behaviors (refer~\cite{Suresh.etal.RAL.2024} for details).
Using this,  Behavioral Entropy $H^B: \Delta_M \times \mathbb{R}_{>0} \to \mathbb{R}_{\geq 0}$ with $\beta = \exp({(1-\alpha)\log(\log(M))})$\footnote{to ensure $H^B$ satisfies the axioms~\cite{Shannon.BellJ.1948, Amigo.etal.Entropy.2018} of generalized entropies, \( \Delta_M \) denotes the space of probability distributions over a finite set of \( M \) outcomes of $X$}, is defined as:
\begin{equation}
\label{eq:behavioral_entropy}
    H^B (X;\alpha) = -\sum_{x\in X}w(p(x))\log(w(p(x)).
\end{equation}

\paragraph*{Expected Information Gain (Mutual Information)}
\label{sec:mutual_info}

Mutual information provides a rigorous measure of the expected reduction in uncertainty and serves as the core metric for evaluating candidate trajectories. Let $X \in \{0,1\}$ represent the belief of existence of hazard at a cell $c \in \mathcal{E}_{\text{grid}}$, and let $Z \in \{0,1\}$ be a binary observation obtained from a deployed sensor or robot. The \textit{expected information gain} from observing $Z$ is quantified by the mutual information:
\begin{equation}
\label{eq:i_bgs}
    I_{BGS}(X;Z) = H(X) - H(X \mid Z),
\end{equation}
where $H(X)$ is the prior entropy over the belief, and $H(X \mid Z)$ is the conditional entropy defined over observation space $Z$. Under the BE framework, we replace classical entropy with its behavioral counterpart:
\begin{equation}
\label{eq:i_B}
    I_B(X;Z, \alpha) = H^B(X; \alpha) - H^B(X \mid Z; \alpha),
\end{equation}

\subsection{Analysis of Behavioral Mutual Information}
\label{sec:theoretical_guarantee}

We now provide a theoretical arguments justifying the use of BE in lieu of classical Boltzmann-Gibbs-Shannon (BGS) entropy for decision making under uncertainty. Specifically, we show that for any binary sensing model and prior belief, there exists a behavioral parameter \( \alpha > 0 \) such that the mutual information under BE exceeds that of SE.

Let \( X \in \{0, 1\} \) be a latent binary variable representing the presence of a hazard, and \( Z \in \{0, 1\} \) be the binary survivability observation with \(\mathbb{P}(Z = 1 \mid X = 1) = \lambda\) (True Positive Rate - TPR), and
\(\mathbb{P}(Z = 1 \mid X = 0) = \gamma\)  (False Positive Rate - FPR). Let \( p = \mathbb{P}(X = 1) \) denote the prior belief. Under the BE framework, this prior is transformed by the Prelec weighting function, 
denoted by \( w(p, \alpha) \).

\begin{theorem}[Existence of an Informative $\alpha$]
\label{thm:informative_alpha}
For any binary observation model satisfying \( 0 < \gamma < \lambda < 1 \) and any prior belief \( p \in (0, 1) \), there exists a positive behavioral parameter \( \alpha > 0 \) such that the mutual information under BE equals or exceeds SE: $I_B(X; Z, \alpha) \ge I_{\mathrm{BGS}}(X; Z)$


\end{theorem}
\vspace{0.1em}
\begin{proof}
The mutual information under classical SE is:
\[
I_{\mathrm{BGS}}(X; Z) = H(\lambda)(1 - p) + H(\gamma) p + H_{\text{obs}}^{\mathrm{BGS}}(Z),
\]
and under BE:
\[
I_B(X; Z, \alpha) = H(\lambda)(1 - w(p)) + H(\gamma) w(p) + H_{\text{obs}}^B(Z;\alpha),
\]
where \( H(\lambda) \) and \( H(\gamma) \) are the binary entropies of the sensor’s false positive and true positive rates, respectively. The entropy shift in the observation distribution is:
\begin{equation}
    \label{eq:delta_i}
    \Delta H_{\text{obs}}(Z) := H_{\text{obs}}^B(Z) - H_{\text{obs}}^{\mathrm{BGS}}(Z),
\end{equation}

\noindent which simplifies the mutual information difference as:
\begin{align}
\Delta I(\alpha) &= I_B(X; Z; \alpha) - I_{\mathrm{BGS}}(X; Z) \nonumber \\
                 &= (H(\gamma) - H(\lambda))(w(p) - p) + \Delta H_{\text{obs}}(Z;\alpha).
\end{align}

\medskip
\noindent
In practice, the product term \( (H(\gamma) - H(\lambda))(w(p) - p) \) is small in magnitude across the belief and parameter space due to two factors: (i) the limited range of $\lambda$ and $\gamma$ entropies \( H(\lambda), H(\gamma) \in [0, \log 2] \), and (ii) the fact that \( w(p) \) remains relatively close to \( p \) for most \(\alpha\). 
Thus, the dominant term influencing the gain in mutual information is the observation entropy shift \( \Delta H_{\text{obs}}(Z) \), which tends to be positive when \( \alpha < 1 \) due to the curvature of the Prelec weighting function.

\medskip
\noindent
Therefore, BE yields greater informativeness than SE if $\Delta H_{\text{obs}}(Z)~>~0$ which holds for $\alpha < 1$.
\end{proof}

\begin{figure}
    \centering
    \begin{subfigure}[t]{0.49\linewidth}
        \centering
        \includegraphics[width=\linewidth]{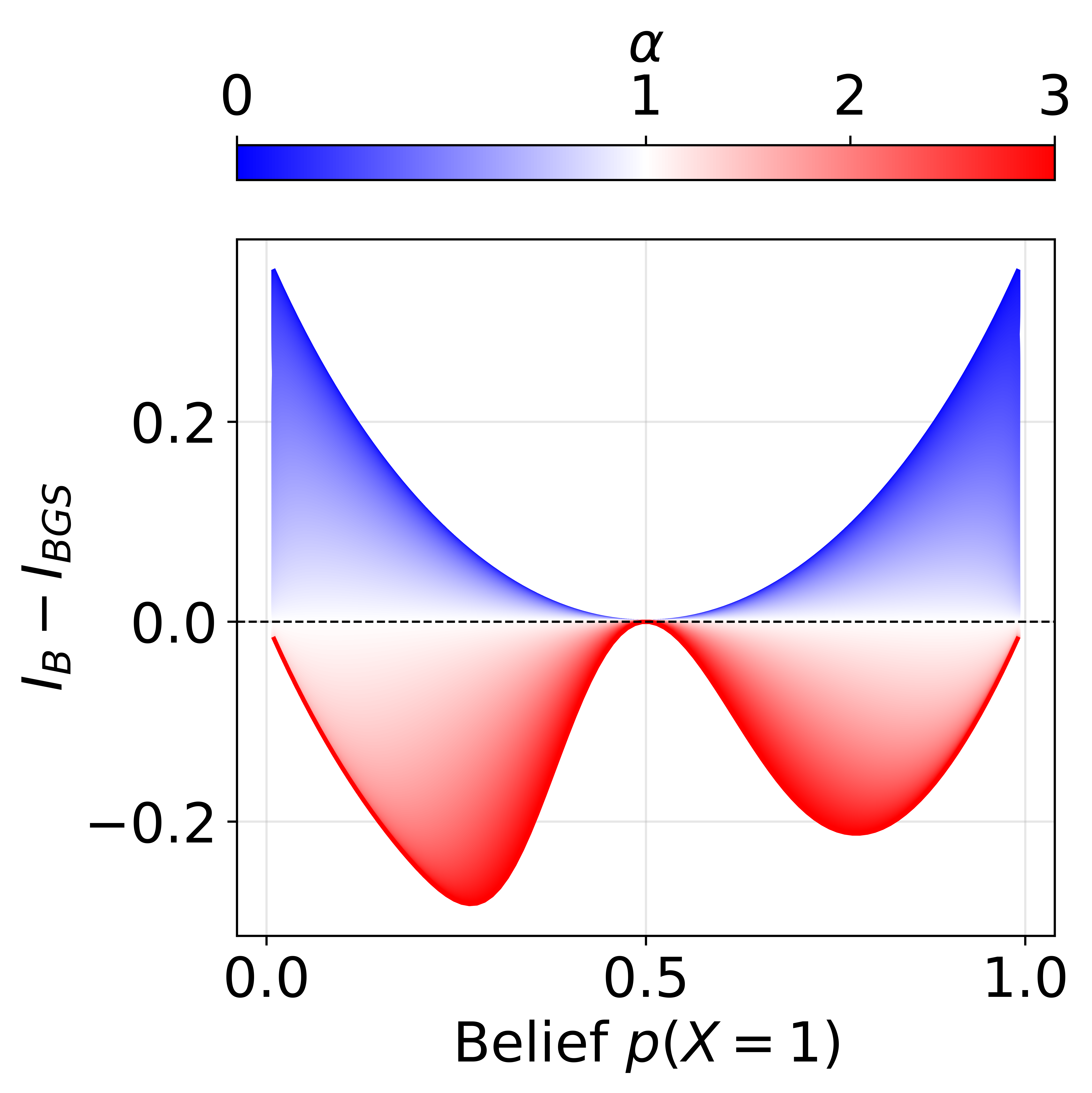}
        \caption{}
        \label{fig:ib_diff_curve}
    \end{subfigure}
    \hfill
    \begin{subfigure}[t]{0.49\linewidth}
        \centering
        \includegraphics[width=\linewidth]{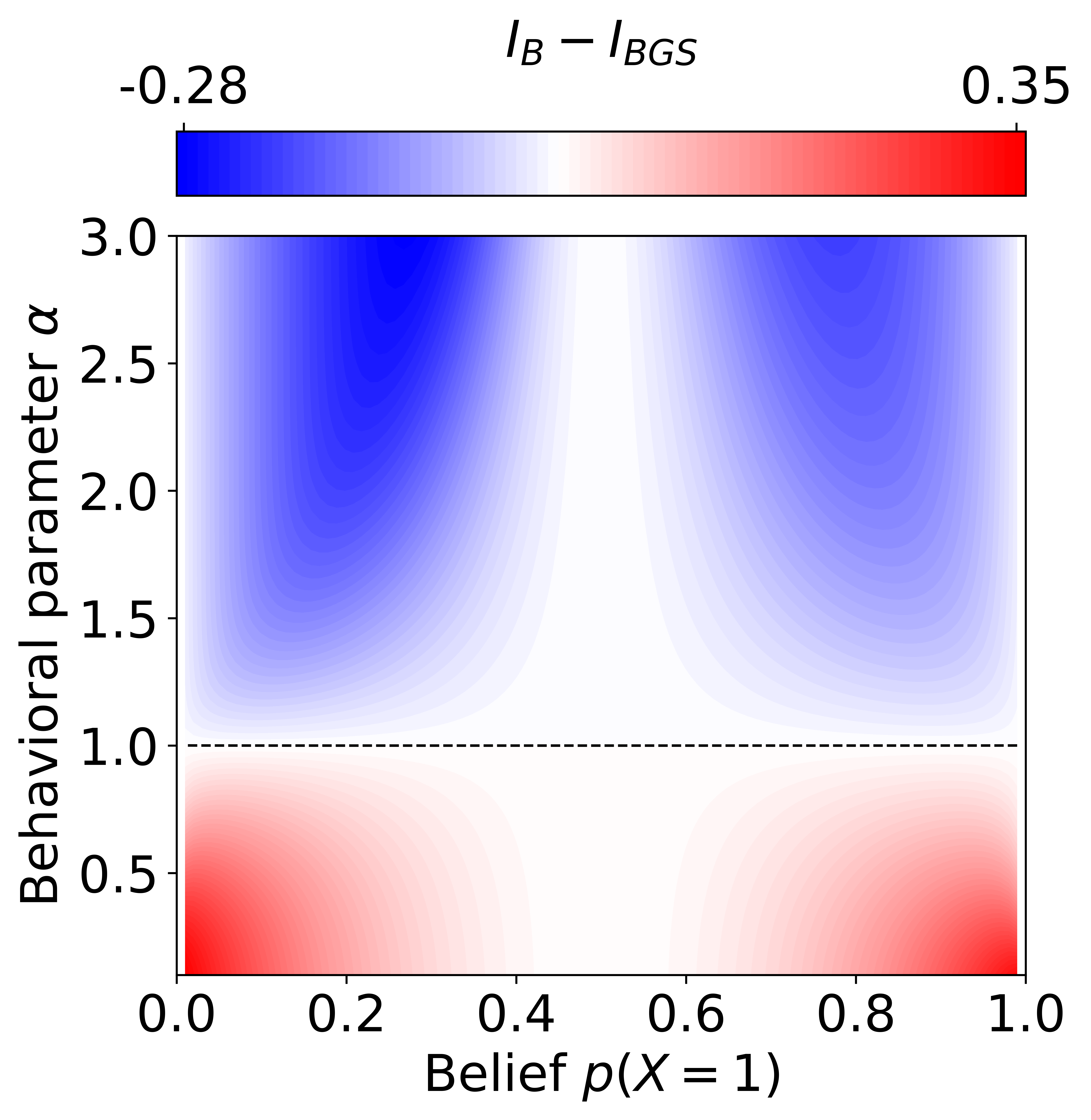}
        \caption{}
        \label{fig:avg_heatmap}
    \end{subfigure}
    \caption{ \small (a) Mutual information difference \( I_B - I_{\text{BGS}} \) across prior beliefs for various \( \alpha \). (b) Averaged gain over all sensor configurations. Dashed contours show where \( I_B = I_{\text{BGS}} \).}
    \vspace{-15pt}
    \label{fig:theoretical_combined}
\end{figure}

\paragraph*{Empirical and Practical Implications}
While Theorem~\ref{thm:informative_alpha} guarantees the existence of an informative $\alpha$, we further validate this behavior empirically. As shown in \Cref{fig:ib_diff_curve}, smaller $\alpha$ values (\( \alpha < 1 \)) often yield higher mutual information \( (I_B - I_{\text{BGS}}) \) across a wide range of prior beliefs. Averaging over realistic sensor parameters \( (\lambda, \gamma) \in [0.70, 0.99] \times [0.01, 0.30] \), the heatmap in \Cref{fig:avg_heatmap} confirms that BE is more diverse than classical entropy. 
These results also reveal an important emergent property. The parameter $\alpha$ acts as a behavior tuning knob that controls the planner's risk sensitivity. This interpretation follows directly from the curvature of the Prelec function \( w(p, \alpha) \), which modifies its subjective belief, resulting in behaviors that can be interpreted along a risk spectrum. We identify three characteristic regimes:

\begin{itemize}
    \item \textbf{Conservative Behavior (\( \alpha < 1 \))}:  
    The transformation exaggerates low-probability events. This leads to \textit{risk-averse} behavior, where the planner tends to explore cautiously, prioritizing safety and wide-area coverage. Hazards are perceived as more imminent than they are.

    \item \textbf{Neutral Behavior (\( \alpha = 1 \))}:  
    The transformation reduces to the identity \( w(p) = p \), recovering classical SE. 

    \item \textbf{Aggressive Behavior (\( \alpha > 1 \))}:  
    The transformation underweights low probabilities and overweights ambiguity near \( p = 0.5 \), causing the planner to focus on high-uncertainty areas. This results in \textit{risk-insensitive} behavior, emphasizing rapid information acquisition at the potential cost of safety by ignoring hazards.
\end{itemize}

This behavioral spectrum naturally emerges from the structure of BE and its effect on information gain as derived above. As such, BE enables principled modulation of robot behavior through a single interpretable parameter. This capability becomes especially important in multi-cell or multi-agent deployments (\Cref{sec:experiments_and_results}), where compounding risk and limited communication demand flexible strategies.

\paragraph*{Trajectory Generation via Behavioral Mutual Information}
We adopt the forward-search trajectory generation procedure from~\cite{otte2021path}, which performs belief rollout over a discrete motion primitive graph to evaluate trajectories of fixed planning horizon \( T \). At each step, feasible actions are expanded using a motion primitive library, generating a set of candidate trajectories. The belief map is then propagated along each path, simulating expected observations to estimate future uncertainty.   In our formulation, we replace the classical Shannon-based mutual information objective with our proposed Behavioral Entropy-based mutual information (see~\cref{sec:mutual_info}). This modification preserves the original framework’s computational structure and anytime behavior, while enabling adaptive risk-aware planning through the tunable behavioral parameter \( \alpha \).

\section{Behavior-Adaptive Path Planning (BAPP) }

We present the \textit{Behavior-Adaptive Path Planning (BAPP)} framework, a unified system for coordinating heterogeneous multi-robot deployments in failure-prone, communication-denied environments. Each robot’s trajectory is generated by maximizing behavior-modulated mutual information, as introduced in~\cref{sec:theoretical_guarantee}, where the parameter $\alpha$ governs risk sensitivity. Built on this foundation, BAPP supports two behavior-adaptive modes: (i) \textbf{BAPP-TID} for intelligent triggering of high-fidelity agents, and (ii) \textbf{BAPP-SIG} for risk-aware, failure-sensitive exploration. A mobile base robot, capable of transporting and relaying data from multiple agents, acts as the central coordinator and periodically relocates to maximize global coverage and information gain.

\subsection{BAPP-TID: Triggered Intelligent Deployment}
\Cref{fig:concept-figure} illustrates the scenario of mapping hazardous sources on unexplored planetary surfaces. In such missions, deploying high-fidelity robots with advanced sensors across the entire environment is often infeasible due to resource constraints. A more scalable alternative is to first deploy low-cost, disposable robots that can rapidly cover ground, accepting that some may fail due to harsh conditions such as radiation, dust, magnetic interference, or unstable terrain. However, these agents are inherently limited in sensing accuracy and resilience. The proposed BAPP-TID framework leverages this cost-performance trade-off by continuously monitoring mission progress and \textit{triggering high-fidelity agents only when exploration begins to stagnate}.

\paragraph*{Entropy-Based Triggering}
Let $H_d$ denote the global map entropy at deployment step $d$. BAPP-TID computes the change in entropy over a sliding window of size $\tau$ as $(\Delta H_d = H_{d - \tau} - H_d)$. If $\Delta H_d$ falls below a threshold, the system infers insufficient progress and triggers a high-fidelity deployment. This behavior is governed by two mission phases:  (i) \textit{Early Opportunistic Triggering}, applied before a deployment step threshold $d^T$, and  (ii) \textit{Late Adaptive Triggering}, applied after $d^T$.
\begin{itemize}
\item \textbf{Phase I – Early Opportunistic Triggering:} For $d \leq d^T$, 
if $\Delta H_d < \theta_{\text{early}}$, a high-fidelity robot is deployed to accelerate initial map understanding.
\item \textbf{Phase II – Late Adaptive Triggering:} For $d > d^T$, the threshold tightens over time using a decaying function \(\epsilon(d) = \max(\epsilon_{\min}, \epsilon_{\max} - \lambda d)\), reflecting increased expectations of progress as the mission matures.
\end{itemize}

\subsection{BAPP-SIG: Safe Information Gathering}

In field deployments, robots are often exposed to hazards such as radiation, fire, or unstable terrain. Aggressive exploration without accounting for these risks can rapidly deplete the available robot pool. BAPP-SIG addresses this by dynamically adapting the BE parameter $\alpha$ in response to the number of robots lost, thus favoring more conservative, safer paths as failures accumulate. 

Let $r_{\text{lost}}$ denote the number of robots lost up to the current deployment, and $r_{\text{total}}$ be the total number of available robots. We define a heuristic for computing the behavioral parameter $\alpha$ that linearly interpolates between predefined bounds:
\begin{equation}
    \alpha = \alpha_{\min} + (\alpha_{\max} - \alpha_{\min}) \cdot \left( \frac{r_{\text{lost}}}{r_{\text{total}}} \right),
\end{equation}
When no robots have failed, $\hat{\alpha} \approx \alpha_{\min}$ encourages conservative behavior. As failures increase, $\hat{\alpha}$ shifts toward $\alpha_{\max}$, gradually allowing more exploratory behavior. While this may seem counterintuitive, early failures are often the only available signal of environmental danger. By cautiously increasing exploration, the system can quickly identify unsafe regions and avoid cascading losses. This adaptive response supports long-term mission preservation.

BAPP-SIG performs a discrete sweep around $\hat{\alpha}$ over a small neighborhood of width $\epsilon$ and selects the path with the highest expected mutual information. This balances exploration and risk in a scalable and mission-aware manner. This process is summarized in~\Cref{alg:bapp-sig}. While it is \textit{theoretically possible} to optimize $\alpha$ over the continuous range $[0, \infty)$ if we want the behavior of the robot to maximize information gain, such optimization is computationally expensive and impractical in real-time settings.



\begin{figure}
    \centering
    \includegraphics[width=0.8\linewidth]{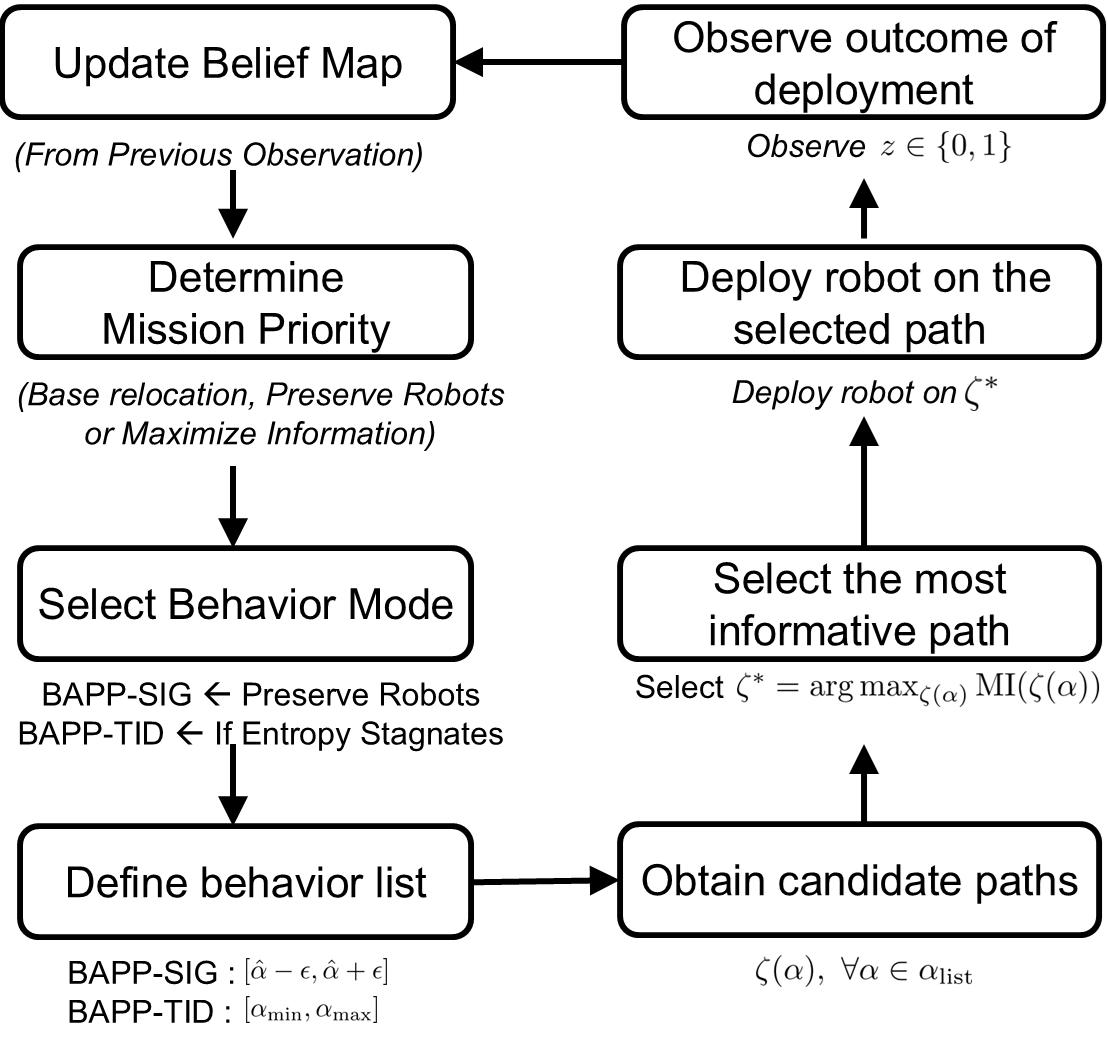} 
    \caption{
        \small
        At each deployment, the belief map is updated based on binary success/failure outcomes. Depending on mission priority behavior is modulated. A discrete set of $\alpha$ values is used to generate candidate paths $\zeta(\alpha)$, and the robot is deployed along the path maximizing mutual information. Observations are then used to update the belief map.
    }
    \label{fig:bapp-pipeline}
\end{figure}

\begin{algorithm}[t]
\caption{BAPP-SIG (Safe Information Gathering)}
\label{alg:bapp-sig}
\begin{algorithmic}[1]
\State \textbf{Input:} Robot loss $r_{\text{lost}}$, total robots $r_{\text{total}}$, graph $\mathcal{G}$, start node $\texttt{startID}$, $\alpha_{\min}, \alpha_{\max}, \epsilon$, step size $\delta$
\State $\hat{\alpha} \gets \alpha_{\min} + (\alpha_{\max} - \alpha_{\min}) \cdot (r_{\text{lost}} / r_{\text{total}})$
\State Initialize $\texttt{I\_list} \gets [\,]$; $\texttt{alphas} \gets \{\hat{\alpha} - \epsilon:\delta:\hat{\alpha} + \epsilon\}$
\ForAll{$\alpha \in \texttt{alphas}$}
    \State $\zeta_\alpha \gets$ generate path from $\texttt{startID}$ using $\alpha$
    \State $I_\alpha \gets$ expected mutual information of $\zeta_\alpha$
    \State Append $I_\alpha$ to $\texttt{I\_list}$
\EndFor
\State $\zeta_{\alpha^*} \gets \zeta_{\arg\max(I_\alpha)}$
\State \textbf{Return:} Path $\zeta_{\alpha^*}$
\end{algorithmic}
\end{algorithm}

\begin{algorithm}[t]
\caption{Entropy-Aware Mobile Base Relocation}
\label{alg:base-relocation}
\begin{algorithmic}[1]
\State \textbf{Input:} Current base $x_b$, belief map $\mathcal{B}$, exploration radius $r_e$, search radius $r_s$, safety threshold $\tau$
\State Initialize candidate set $\mathcal{C} \gets \emptyset$
\ForAll{$\delta \in [-r_s, r_s]^2$}
    \State $x' \gets x_b + \delta$
        \If{\text{\small \Call{IsReachableSafely}{$x_b$, $x'$}} \textbf{and} $\mathcal{B}(x') < \tau$}

        \State Simulate robot partitioning at $x'$
        \State Compute regional entropy within $r_e$

        \State Add $(x', \texttt{mean\_entropy})$ to $\mathcal{C}$
\EndIf \EndFor
\If{$\mathcal{C}$ not empty}
    \State $x^* \gets \arg\max_{(x',h) \in \mathcal{C}} h$
    \Comment{Move base to $x^*$}
\EndIf
\end{algorithmic}
\end{algorithm}

\subsection{Entropy-Aware Mobile Base Relocation and Partitioning}
The mobile base robot plays a dual role in the BAPP framework: (i) it serves as the sole communication and coordination hub in a communication-denied environment, and (ii) periodically relocates to maximize global information gain. Instead of remaining static, it selects safe, informative sites that support effective spatial partitioning. The partitioning is crucial for enabling distributed, non-overlapping exploration without real-time communication. By simulating partition-aware entropy, the base optimizes its position to reduce redundancy and enhance mission efficiency and resilience. 

For each region $\mathcal{R}_i$, the mean entropy is computed as:
\[
H(\mathcal{R}_i) = \frac{1}{|\mathcal{R}_i \cap \mathcal{B}(x', r_e)|} \sum_{p \in \mathcal{R}_i \cap \mathcal{B}(x', r_e)} H(\mathcal{B}(p)),
\]
where $\mathcal{B}(p)$ denotes the belief of adversary presence at point $p$, and $H(\mathcal{B}(p))$ is the binary entropy at that location. The set $\mathcal{B}(x', r_e)$ represents a ball of radius $r_e$ around candidate location $x'$. The candidate’s score is the average entropy in all regions, and the base relocates to the point with the highest such score, allowing coordinated and decentralized deployments in high-uncertain zones.

\paragraph*{BAPP Framework Summary}The entire BAPP pipeline is summarized in~\Cref{fig:bapp-pipeline}, which depicts how belief updates, mission priorities, behavior modulation, and deployment decisions are orchestrated in a closed-loop process. Each cycle begins with an updated belief map and ends with a robot deployment, incorporating entropy-triggered behaviors (BAPP-TID) and survivability-aware planning (BAPP-SIG), along with mobile base relocation when necessary. This structured flow enables robust, risk-sensitive exploration under uncertainty and failure.

\section{Experiments and Results}
\label{sec:experiments_and_results}
\begin{figure*}
    \centering

    \begin{subfigure}[t]{\textwidth}
        \centering
        \includegraphics[width=\textwidth]{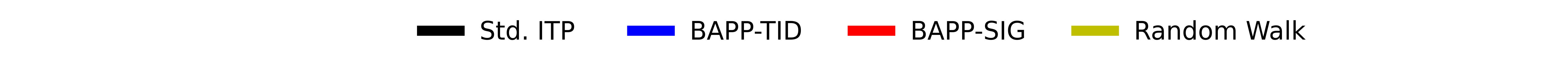}
    \end{subfigure}

    \begin{subfigure}[t]{0.48\linewidth}
        \centering
        \includegraphics[width=\linewidth]{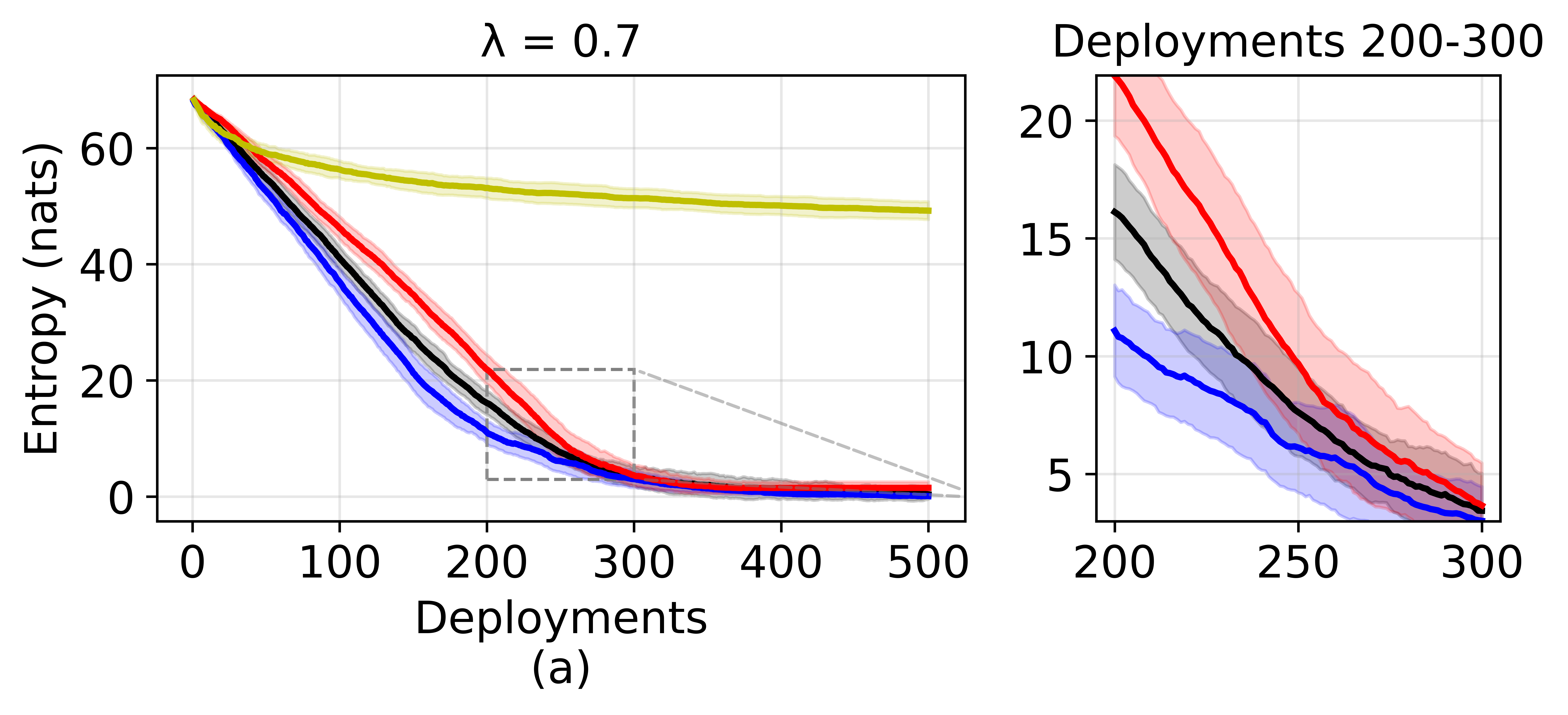}
    \end{subfigure}
    \hfill
    \begin{subfigure}[t]{0.48\linewidth}
        \centering
        \includegraphics[width=\linewidth]{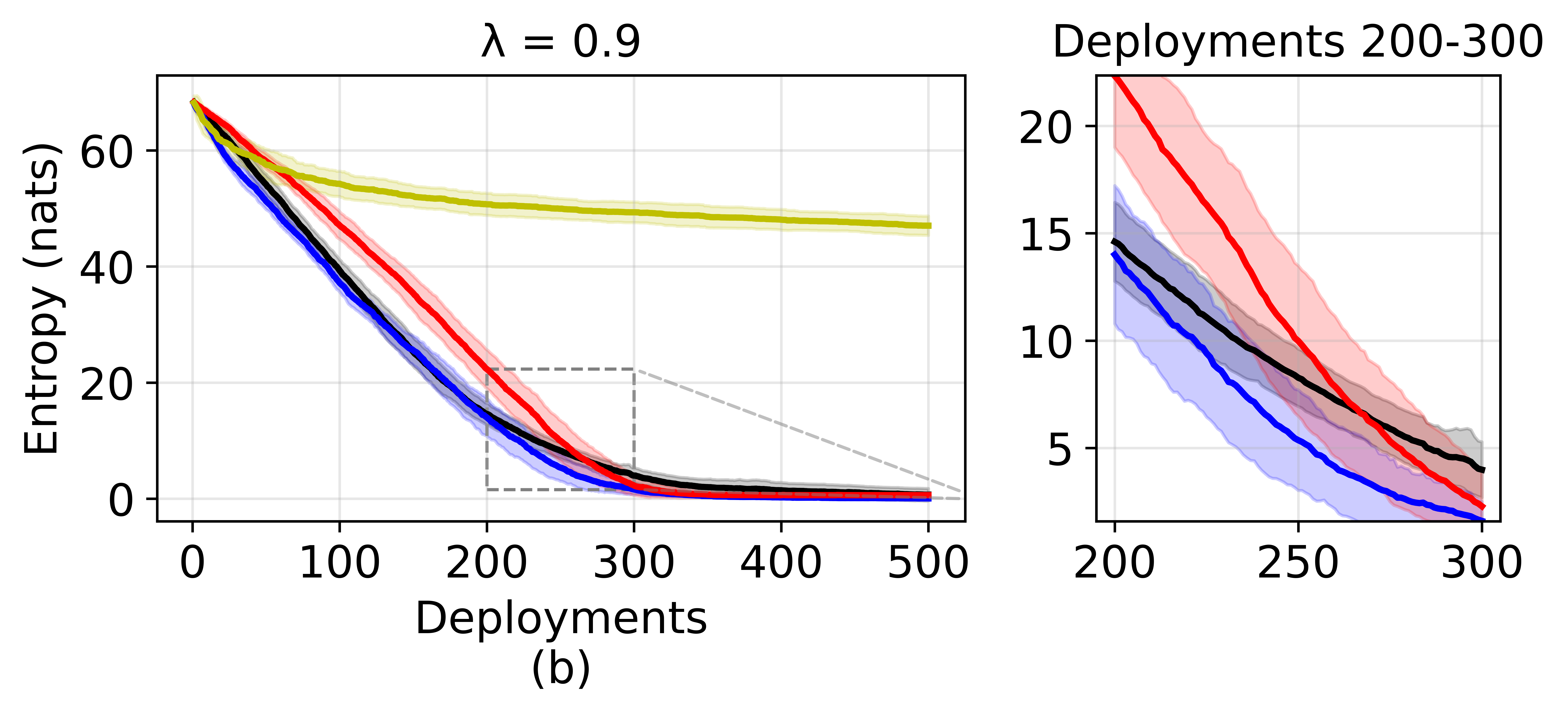}
    \end{subfigure}

    \begin{subfigure}[t]{0.48\linewidth}
        \centering
        \includegraphics[width=\linewidth]{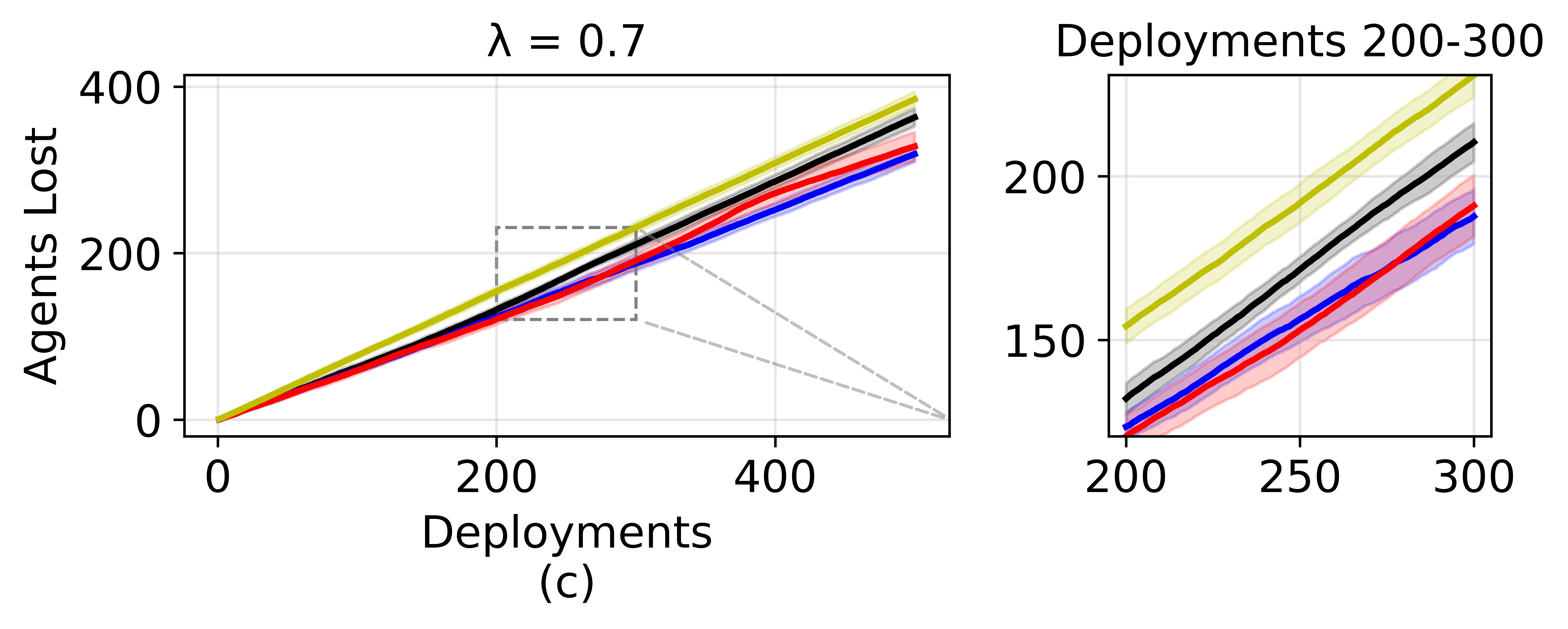}
    \end{subfigure}
    \hfill
    \begin{subfigure}[t]{0.48\linewidth}
        \centering
        \includegraphics[width=\linewidth]{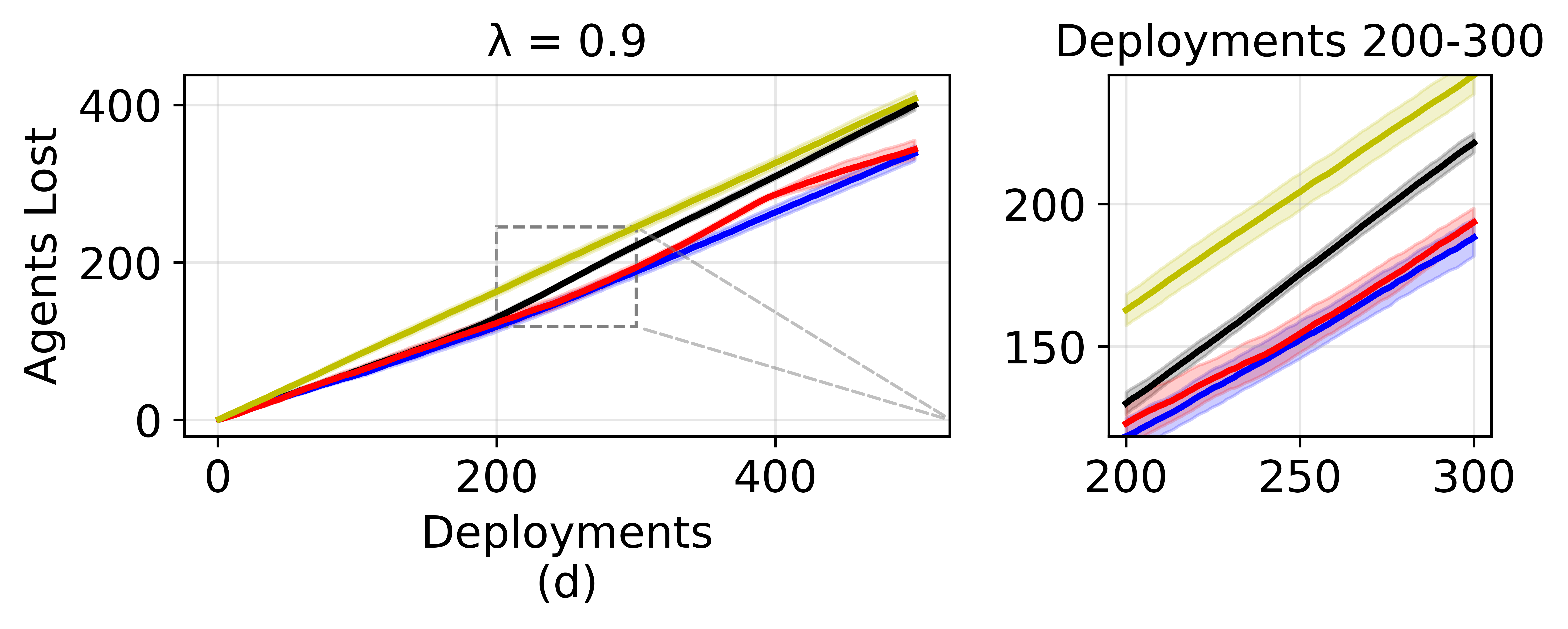}
    \end{subfigure}

    \caption{
        \small (a–b) \textit{Entropy reduction} under moderate ($\lambda = 0.7$) and high ($\lambda = 0.9$) hazard lethality. 
        (c–d) \textit{Cumulative agent loss} under the same conditions. 
        BAPP-SIG improves survivability while BAPP-TID accelerates uncertainty reduction. 
    }
    \vspace{-15pt}
    \label{fig:proof-concept-results}
\end{figure*}

\begin{figure}
    \centering

    \begin{subfigure}[t]{\columnwidth}
        \centering
        \includegraphics[width=\columnwidth]{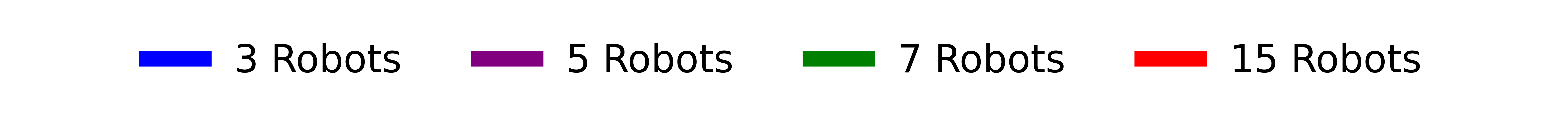}
    \end{subfigure}

    \begin{subfigure}[t]{0.49\linewidth}
        \centering
        \includegraphics[width=\linewidth]{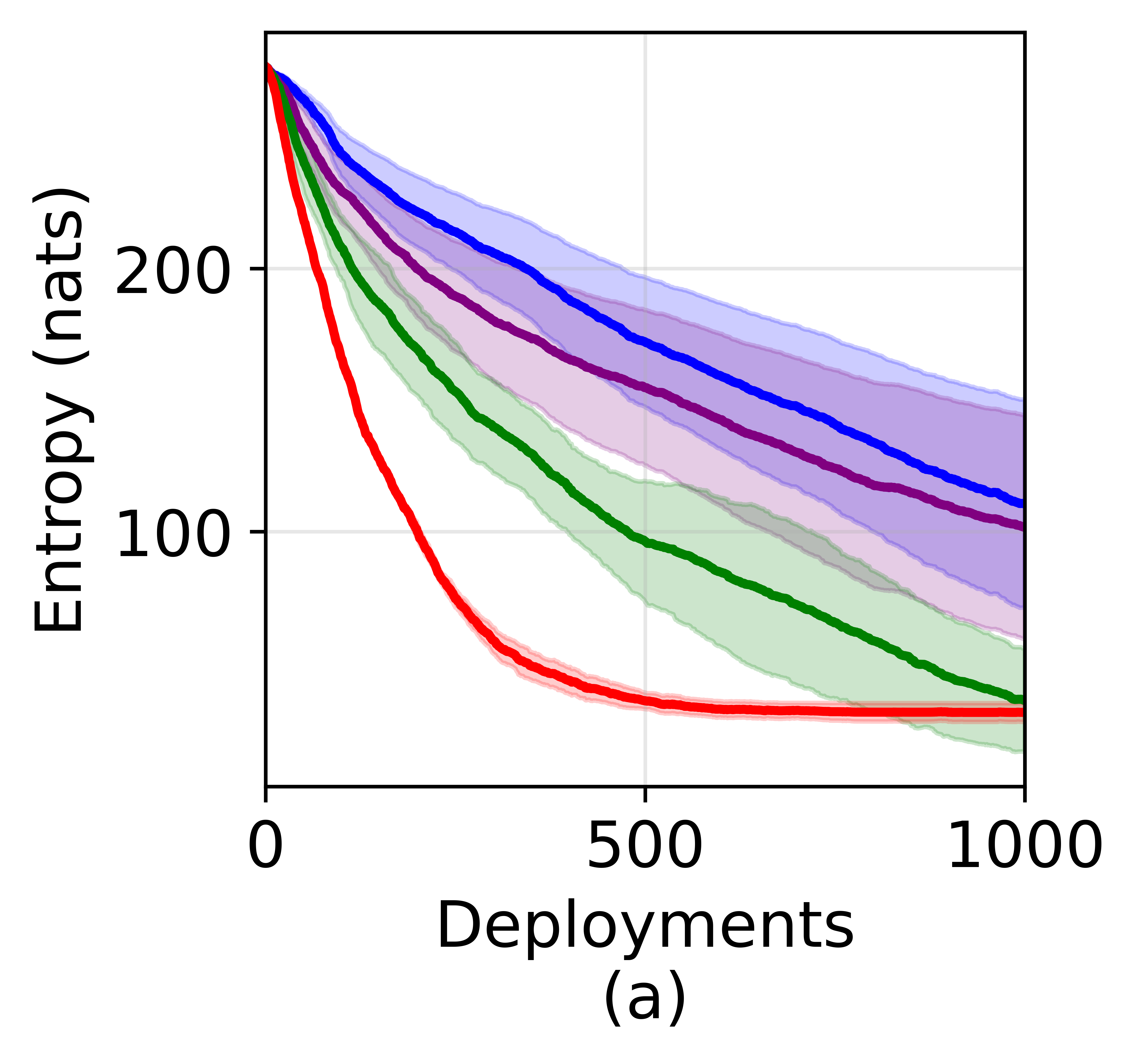}
    \end{subfigure}
    \hfill
    \begin{subfigure}[t]{0.49\linewidth}
        \centering
        \includegraphics[width=\linewidth]{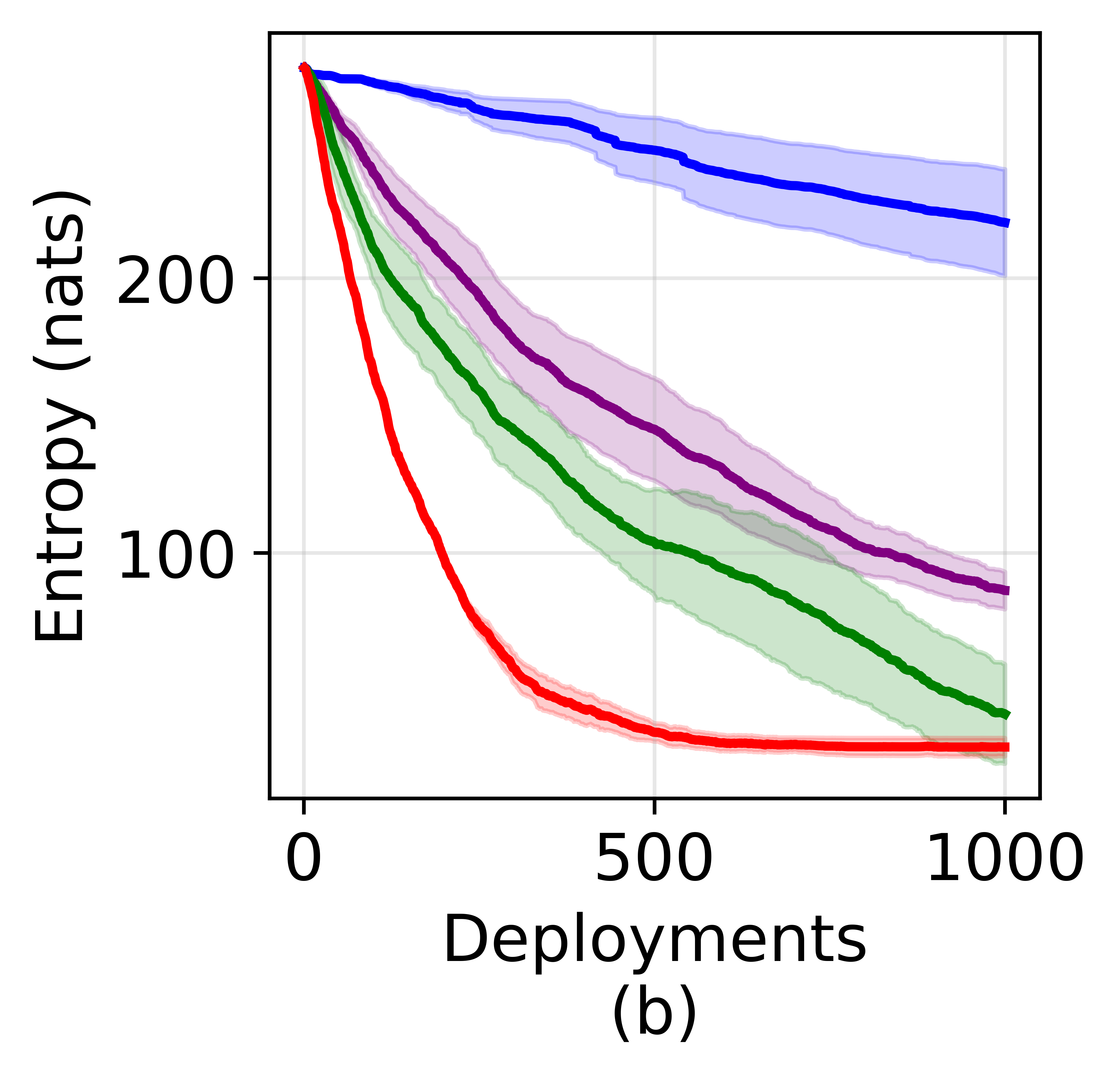}
    \end{subfigure}

    \caption{\small 
        Multi-robot deployment performance. 
        (a) Increasing the number of robots accelerates entropy reduction. 
        (b) Under a fixed energy budget ($n \cdot T = 105$), distributing effort across more, shorter trajectories improves robustness and planning efficiency relative to fewer, longer paths.}
    \vspace{-15pt}
    \label{fig:multirobot-results}
\end{figure}

We evaluate the effectiveness of the BAPP framework in hazardous, communication-denied environments using both single-agent and multi-robot simulations. Our simulations are designed to assess how BAPP improves exploration efficiency, robustness to failure, and scalability under varying levels of environmental risk. We first validate behavior modulation in controlled proof-of-concept single-agent experiments, followed by multi-robot experiments that test coordination, scalability, and energy efficiency.

\subsection{Proof-of-Concept: Validating Behavioral Adaptation}
\label{subsec:proof-of-concept}

In this experiment, we validate the theoretical insight from \Cref{sec:theoretical_guarantee} in a realistic multi-cell, path-based planning setting using a $10 \times 10$ discrete environment.

\paragraph*{Experimental Setup.} We conduct 25 Monte Carlo trials simulating single-agent exploration in a $10 \times 10$ grid. Each agent follows a 9-connected trajectory of fixed length $T = 15$. Cells have hazard lethality $\lambda \in \{0.7, 0.9\}$, and agents incur a 10\% per-step failure probability to emulate adverse conditions (e.g., heat or radiation). The belief map is initialized uniformly and updated after each deployment using binary success/failure feedback.

\paragraph*{Results} We compare our behavior-adaptive planners—\textit{BAPP-SIG} and \textit{BAPP-TID}—against two baselines: a classical Shannon-based planner (\textit{Std-ITP})~\cite{otte2021path, srivastava2022distributed} and a \textit{Random Walk} strategy. \Cref{fig:proof-concept-results} shows planner performance under moderate ($\lambda = 0.7$) and high ($\lambda = 0.9$) hazard lethality. We evaluate each method using two metrics: (i) entropy reduction over deployments, measuring information gain, and (ii) cumulative agent loss, capturing mission survivability.

BAPP-TID consistently achieves the fastest entropy reduction, validating the effectiveness of its trigger-based deployment strategy in accelerating map understanding. BAPP-SIG, while slightly slower in information gain, minimizes agent loss by favoring safer paths and is more computationally efficient, performing a narrower $\alpha$-sweep compared to BAPP-TID's broader search over larger uncertainty thresholds. Both BAPP variants outperform the \textit{Std-ITP} and \textit{Random Walk} baselines. Overall, these results confirm that behavioral modulation in BAPP allows planners to adapt exploration strategies to mission needs, balancing risk and reward in complex, uncertain environments. Multi-robot extensions are presented next in \Cref{sec:multirobot}.

\subsection{Scalable Multi-robot Deployment}
\label{sec:multirobot}

Building on the single-agent results (\Cref{subsec:proof-of-concept}), we evaluate how BAPP-TID scales in multi-robot deployments. By leveraging spatial partitioning and robot isolation, we can render the global exploration problem equivalent to running $n$ parallel single-robot missions. This design principle eliminates inter-robot interference and ensures that the advantages of the single robot BAPP variants are directly carried over to the distributed setting. Therefore, we focus solely on comparing the performance of our BAPP stack across different deployment architectures and coordination schemes, rather than re-evaluating standard entropy-based planners or random walks.

\textbf{\emph{Experimental Setup}}:
We simulate heterogeneous teams in a $20 \times 20$ grid with hazard lethality $\lambda = 0.9$. The team consists of: (i) \textit{disposable explorers}—low-cost, failure-prone robots with a 10\% per-step malfunction rate; (ii) \textit{high-fidelity robots}—sensor-rich, expensive agents with 1\% per-step malfunction rate; and (iii) a \textit{mobile base}, which transports agents, partitions space, and relocates to high-entropy regions. At each relocation step, the base performs radial partitioning of the map and assigns disjoint regions to robots, supporting communication-free deployment. 

We conduct two experimental evaluations:
\begin{itemize}
    \item \textbf{Scalability:} We vary the team size $n \in \{3, 5, 7, 15\}$ while keeping the path length fixed at $T = 15$ to assess how team size impact entropy reduction and mission robustness. We do not relocate the mobile base.

    \item \textbf{Energy Efficiency:} We fix the total energy budget to $n \cdot T = 105$ and compare different team configurations: shallow deployments such as $15 \times 7$, $7 \times 15$, and versus a deeper deployment of $5 \times 21$  and $3 \times 35$. This evaluation highlights the trade-offs between wider distributed coverage and deeper individual exploration.
\end{itemize}

\textbf{\emph{Results}}: 
\Cref{fig:multirobot-results}-(a) shows that increasing the number of deployed robots from $n=3$ to $n=15$ (with $T=15$) accelerates entropy reduction and improves robustness. The 15-robot team reduces entropy to 50\% approximately 68.06\% faster than the 3-robot case. Notably, the entropy curves for $n=3$ and $n=5$ exhibit higher variance, which arises from incomplete spatial coverage, i.e., fewer robots mean the mobile base may not be positioned to support full-area partitioning, resulting in uneven exploration. 

In \Cref{fig:multirobot-results}-(b), we observe that under a fixed energy budget ($n \cdot T = 105$), the $15 \times 7$ configuration achieves the fastest and deepest entropy reduction, outperforming the $3 \times 35$, $5 \times 21$, and $7 \times 15$ setups. Notably, increasing the number of agents while shortening individual path lengths consistently improves exploration efficiency. This confirms that distributing effort across more, shorter trajectories reduces per-agent exposure and enhances spatial coverage in high-risk environments. Furthermore, shorter paths reduce the per-deployment computational burden, yielding additional benefits in real-time energy-constrained planning.

\section{Conclusion}
We present a behavior-adaptive path planning framework for hazard mapping in failure-prone, communication-denied environments. Our approach enables heterogeneous multi-robot teams to dynamically modulate exploration strategies through a tunable behaviors, balancing risk sensitivity and information gain. We theoretically demonstrate the informativeness of BE over classical SE in such environments and validate our findings through both single-agent and multi-agent simulations. Our modular BAPP stack, with its role-aware deployment modes, consistently outperforms traditional planners in both robustness and efficiency.
In future, we plan to incorporate actuation uncertainties into the belief update and path planning, and extend the observation model beyond binary feedback to include partial or probabilistic sensing, enabling more nuanced reasoning under uncertainty. 
\bibliographystyle{IEEEtran}
\bibliography{refs}

\end{document}